\documentclass{article}

\PassOptionsToPackage{numbers, compress}{natbib}
\usepackage{colortbl}
\usepackage[dvipsnames]{xcolor}
\usepackage[final,main]{robotwin_2025}
\usepackage{marvosym}

\usepackage{wrapfig}

\usepackage[utf8]{inputenc} 
\usepackage[T1]{fontenc}    
\usepackage{pifont}
\usepackage{hyperref}       
\usepackage{url}            
\usepackage{booktabs}       
\usepackage{amsfonts}       
\usepackage{nicefrac}       
\usepackage{microtype}      
\usepackage{xcolor}         
\usepackage{listings}
\usepackage{graphicx}
\usepackage{array}
\usepackage{amsmath}
\usepackage{amssymb}
\usepackage{adjustbox}
\usepackage{tcolorbox}
\usepackage{multirow}
\usepackage{makecell} 

\definecolor{robotwincolor1}{HTML}{65a487}
\definecolor{robotwincolor2}{HTML}{68349a}

\title{
AnchorDP3: 3D Affordance Guided Sparse Diffusion Policy for Robotic Manipulation \\ \vspace{1em}
\normalfont \large \textit{RoboTwin Challenge Simulation Champion @ CVPR 2025 MEIS Workshop} \\
\normalfont{\textcolor{robotwincolor1}{Robo}\textcolor{robotwincolor2}{Twin} Dual-Arm Collaboration Challenge Technical Report} \\
}

\author{
    JD-TFS Team \\
    Ziyan Zhao\textsuperscript{1,2} \hspace{1em}
    Ke Fan\textsuperscript{3} \hspace{1em}
    He-Yang Xu\textsuperscript{4} \hspace{1em}
    Ning Qiao\textsuperscript{1} \\
    Bo Peng\textsuperscript{1,5}  \hspace{1em}
    Wenlong Gao\textsuperscript{1}  \hspace{1em}
    Dongjiang Li\textsuperscript{1}  \hspace{1em}
    Hui Shen\textsuperscript{1} \\
    \small \textsuperscript{1}Jingdong Technology Information Technology Co., Ltd \hspace{1em}
    \small \textsuperscript{2}Tsinghua University \\
    \small \textsuperscript{3}Fudan University \hspace{1em}
    \small \textsuperscript{4}Southeast University \hspace{1em}
    \small \textsuperscript{5}University of Science and Technology of China
}

\begin{document}

\maketitle

\begin{abstract}
We present \textbf{AnchorDP3}, a diffusion policy framework for dual-arm robotic manipulation that achieves state-of-the-art performance in highly randomized environments. AnchorDP3 integrates three key innovations: 1) \textbf{Simulator-Supervised Semantic Segmentation}, using rendered ground truth to explicitly segment task-critical objects within the point cloud, which provides strong affordance priors; 2) \textbf{Task-Conditioned Feature Encoders}, lightweight modules processing augmented point clouds per task, enabling efficient multi-task learning through a shared diffusion-based action expert; 3) \textbf{Affordance-Anchored Keypose Diffusion with Full State Supervision}, replacing dense trajectory prediction with sparse, geometrically meaningful action anchors, i.e., keyposes such as pre-grasp pose, grasp pose directly anchored to affordances, drastically simplifying the prediction space; the action expert is forced to predict both robot joint angles and end-effector poses simultaneously, which exploits geometric consistency to accelerate convergence and boost accuracy. Trained on large-scale, procedurally generated simulation data, AnchorDP3 achieves a \textbf{98.7\% average success rate} in the RoboTwin benchmark across diverse tasks under extreme randomization of objects, clutter, table height, lighting, and backgrounds. This framework, when integrated with the RoboTwin real2sim pipeline, has the potential to enable fully autonomous generation of deployable visuomotor policies from only scene and instruction—totally eliminating human demonstrations from learning manipulation skills.
\end{abstract}

\section{Introduction}

Robotic manipulation in unstructured, real-world environments remains a longstanding challenge, particularly when tasks require dual-arm coordination, tool use, and generalization across diverse object categories and scene layouts. Recent simulation benchmarks, such as RoboTwin~\cite{robotwin} and Mimicgen~\cite{mimicgen}, aim to accelerate progress in this area by providing large-scale, procedurally generated environments along with automated task definitions and expert data synthesis. These platforms offer a promising foundation for training visuomotor policies entirely in simulation, enabling fast iteration and reduced reliance on costly real-world demonstrations.

\begin{wrapfigure}[18]{r}{0.6\textwidth}
\vspace{-1em}
\centering
\includegraphics[width=0.58\textwidth]{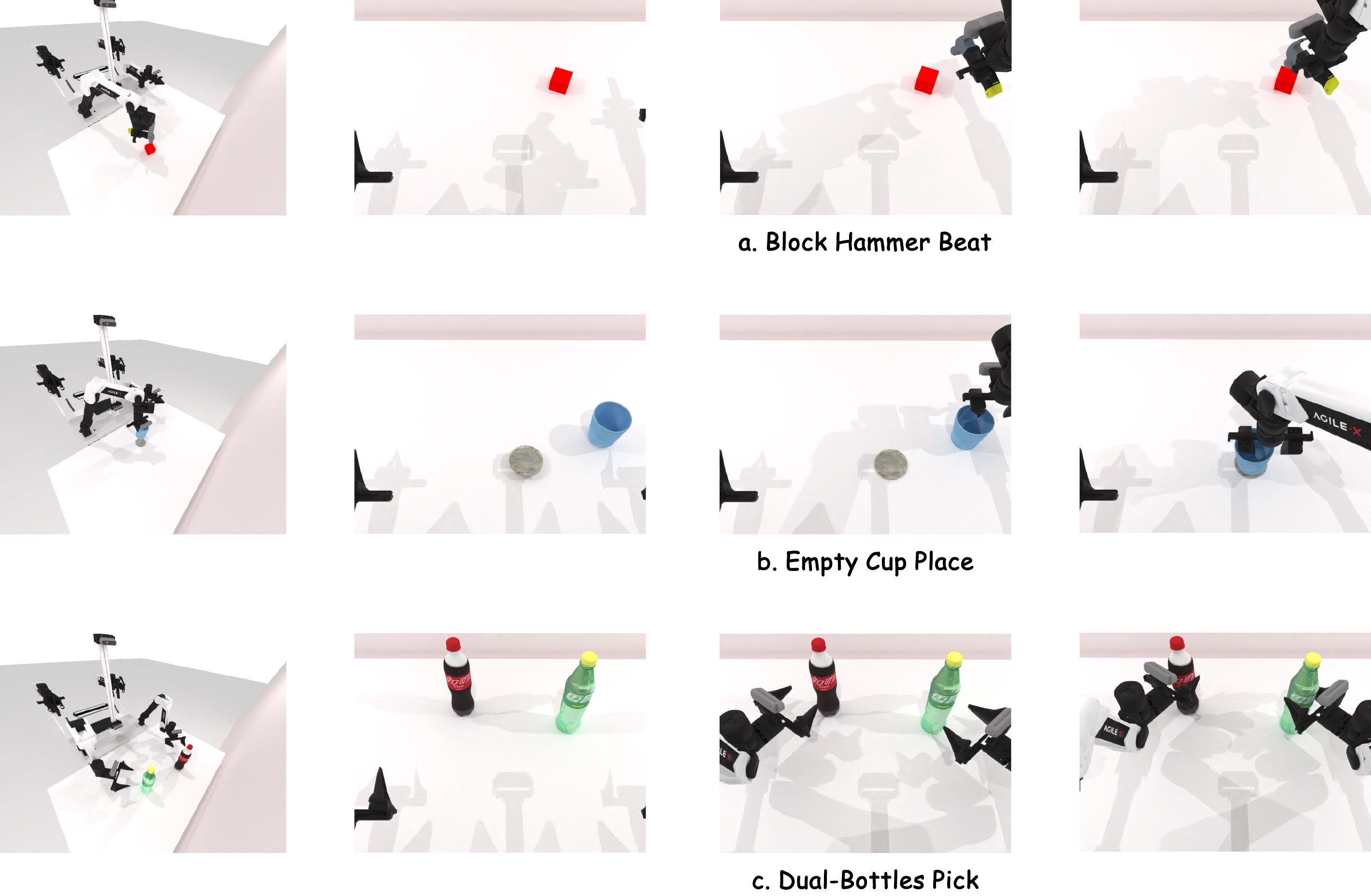}
\caption{\small Examples of 3D simulation tasks generated by the RoboTwin system, including diverse object configurations and manipulation goals across multiple environments.}
\label{fig:data}
\end{wrapfigure}

As part of the RoboTwin Dual-Arm Collaboration Challenge\footnote{\url{https://robotwin-platform.github.io/}}, which centers on learning dual-arm visuomotor policies from simulation alone, participants are tasked with developing a single end-to-end model that can follow language instructions and operate under extreme scene randomization. The RoboTwin pipeline automatically constructs diverse simulation environments by generating 3D assets from RGB images using AIGC techniques, assigning affordance coordinates via 3D semantic models, and synthesizing expert trajectories with large language models. Figure~\ref{fig:data} illustrates examples of diverse 3D simulation scenes generated by the RoboTwin system, covering a range of manipulation tasks. Despite these advantages, our participation in the RoboTwin Dual-Arm Collaboration Challenge revealed several key limitations that hinder the development of scalable, generalizable end-to-end policies. In particular, we identified three core challenges: perceptual ambiguity in cluttered scenes, task interference under shared representation learning, and inefficient action modeling for long-horizon, multi-step tasks.


To address these challenges, we propose AnchorDP3, a diffusion-policy-based framework for dual-arm robotic manipulation under high scene variability. It comprises three key components, each aligned with specific bottlenecks identified in the RoboTwin Challenge.
(1) Simulator-Supervised Semantic Segmentation improves perception in cluttered environments by training a point-level segmentation model using simulation-rendered object-part masks. This enables the policy to prioritize task-relevant regions in the point cloud with minimal ambiguity.
(2) Task-Conditioned Feature Encoders mitigate multi-task interference by introducing lightweight, per-task modules that extract specialized point cloud features. These features feed into a shared diffusion-based action expert, supporting scalable multi-task learning without degrading individual task performance.
(3) Affordance-Anchored Keypose Diffusion with Full State Supervision addresses the inefficiency of dense trajectory modeling by predicting sparse, geometrically anchored keyposes (e.g., pre-grasp, grasp). Joint supervision on joint angles and end-effector poses ensures geometric consistency and improves learning stability.
The resulting policy achieves a 98.7\% average success rate across diverse tasks, demonstrating strong generalization under severe randomization.

AnchorDP3 was evaluated as part of the RoboTwin Dual-Arm Collaboration Challenge, where it achieved the top performance in both simulation tracks I and II of the competition. These results highlight the practical advantages of our framework under diverse, randomized conditions, and validate its potential as a scalable solution for dual-arm visuomotor policy learning in simulation. Our main contributions are summarized as follows:
\begin{itemize}
\item We propose AnchorDP3, a modular diffusion policy framework for dual-arm robotic manipulation, designed for high generalization in procedurally randomized environments.
\item We leverage simulation-rendered semantics to build a lightweight perception pipeline that injects affordance-awareness into point cloud representations.

\item We introduce task-conditioned encoding and affordance-aligned keypose supervision to resolve multi-task interference and improve long-horizon action efficiency.

\item Our approach attained the highest success rate (98.7\%) in the RoboTwin Challenge simulation tracks, validating the potential of fully simulated training for complex dual-arm manipulation tasks.
\end{itemize}

\section{Related Work}
\subsection{Visuomotor policy learning based on diffusion models}

Recently, diffusion models~\cite{ho2020denoising, rombach2022high} have emerged as a powerful new tool for robot learning. Diffusion Policy~\cite{chi2023diffusion} frames the robot vision-to-action policy as a conditional denoising diffusion process, boosting success rates across several manipulation benchmarks while naturally accommodating multimodal distributions and high-dimensional action spaces. 3D  diffusion policy~\cite{ze20243d} integrates compact 3-D point-cloud representations into diffusion policies, enabling data-efficient and highly generalizable robot imitation.
DPPO~\cite{ren2024diffusion} presents a policy-gradient fine-tuning framework for diffusion-based policies that exploits their structure to enable efficient and robust learning. Equivariant DP~\cite{wang2024equivariant} embeds SO(2) symmetry into diffusion-based behavior cloning, enabling a denoising function that generalizes well, especially on simulated and real 6-DoF manipulation tasks.

\subsection{Affordance-guided motion generation}
The concept of "affordance", initially introduced by Gibson \cite{gibsonafford}, refers to the possibilities for action that the environment provides to an individual. Originating from the 2D domain, initial work in affordance detection primarily focused on identifying objects with affordances \cite{affordnet}. Subsequent approaches incorporated linguistic information to enhance detection performance \cite{lu2022phrase}, but still operated at a coarse object level without capturing finer affordance structures. To address this, more recent efforts~\cite{afford1,afford2,afford3,afford4} shifted attention toward part-level affordance localization, significantly improving detection granularity. As embodied AI progressed, affordance learning expanded into the 3D domain. For instance, 3D AffordanceNet~\cite{afford5} introduced one of the earliest datasets enabling affordance understanding from object point clouds, while IAGNet~\cite{afford6} proposed image-query-driven 3D part affordance detection. More recently, open-vocabulary affordance understanding in 3D has emerged~\cite{afford7}. Unlike prior work, we explicitly extract affordance cues from simulator-rendered layers to train a semantic segmentation model, append affordance features to point clouds, and anchor keypose prediction to affordance-aligned constraints, effectively simplifying the output space. However, there is hard work that focuses on improving the performance of the diffusion policy algorithm based on the keyposes. 

\subsection{Multi-task learning for robotic manipulation}

Recent work demonstrates that leveraging large-scale, diverse datasets can significantly improve multi-task learning for robotic manipulation. RT-1~\cite{brohan2022rt} embeds instructions with the Universal Sentence Encoder and uses FiLM layers to enable robots to perform various tasks. GR-2 ~\cite{cheang2024gr} employs a GPT-style visual manipulation model, pretrained on video data and fine-tuned on robot data, to jointly predict future frames and action sequences.

The $\pi_0$~\cite{black2410pi0} model utilizes a vision-language backbone pretrained on internet-scale image-text pairs, combined with an action expert, to enable effective language-driven multi-task learning. RoboVLMs~\cite{li2024towards} systematically analyze design choices for vision-language-action models, providing insights for multi-task settings. FAST~\cite{pertsch2025fast} introduces a tokenizer that converts continuous robot actions into discrete tokens, allowing VLA models like Transformers to handle complex robotic behaviors more efficiently.

\section{AnchorDP3 Framework}

\subsection{Overview}

\begin{figure}[h!]
    \centering
    \hspace*{-0.05\linewidth} 
    \includegraphics[width=1.1\linewidth]{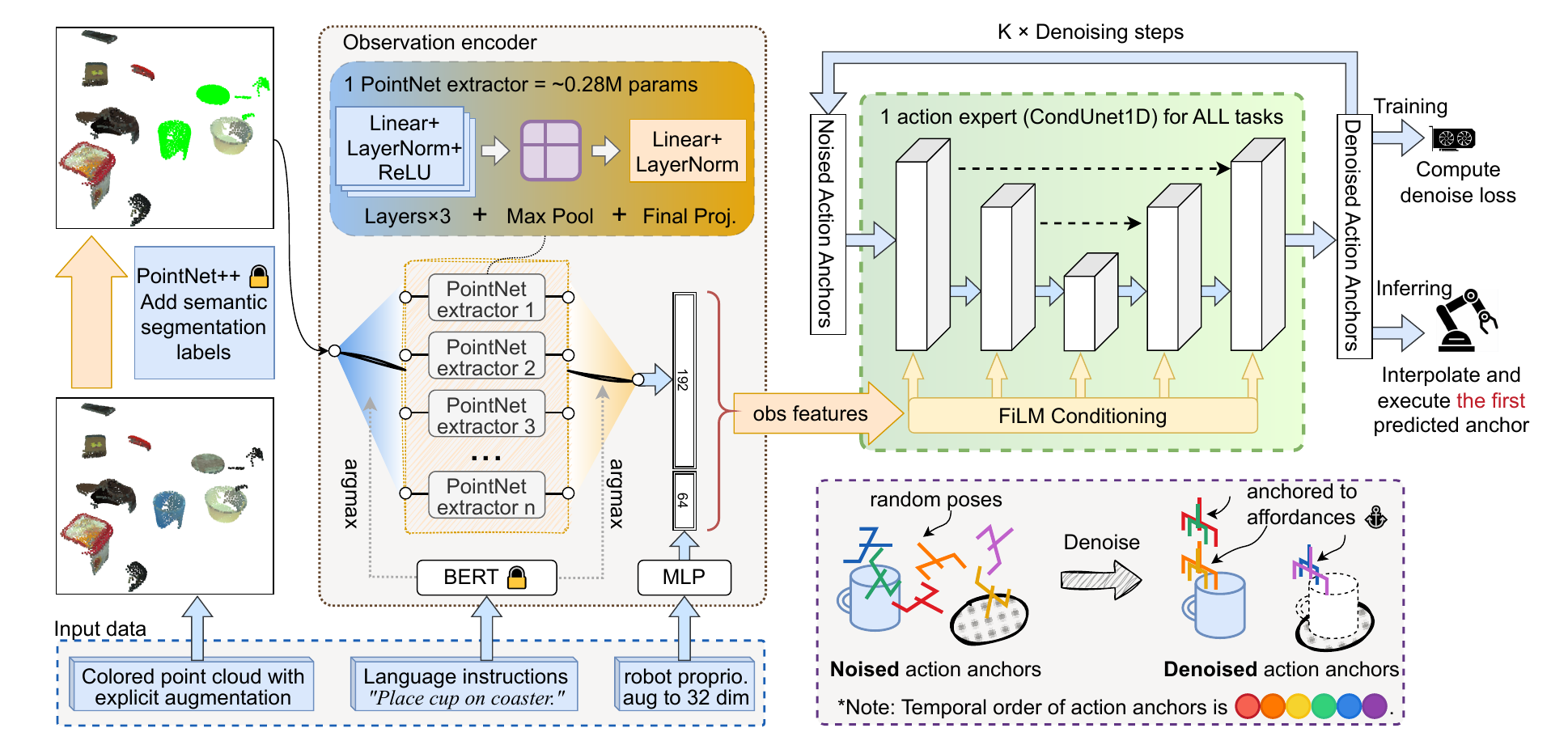}
    \caption{The overall workflow of AnchorDP3.}
    \label{fig:workflow}
    \vskip -0.15in
\end{figure}

AnchorDP3 is a diffusion-based policy framework designed for vision-language-conditioned robotic manipulation in randomized environments. It processes RGB-D observations and language instructions to predict sparse keypose sequences anchored to 3D affordances. The overall workflow of AnchorDP3 is shown in Figure~\ref{fig:workflow}.

\subsection{Observation processing}

Preprocessing and augmenting observations can reduce the difficulty of neural network learning. We employed several methods to significantly augment state, action, and RGBD observations. Our strategy extends the dimension of proprioception and overall adopts a method based on 3D point cloud observation.

\textbf{32-dim robot state and action.} In this work, the robot state and action are augmented to 32 dimensions, as listed in Table~\ref{table:dim}.

We use both the joint space vector (qpos) and end effector pose to represent the robot state and action. Note that the rotation is not represented through Euler Angles or quaternions. Instead, we use the first two columns of the rotation matrix ${R^{3\times3}}$ to construct a continuous representation of rotation, which is beneficial for training neural networks \cite{zhou2019continuity}.

\begin{table}[h]
\centering
\caption{State and Action Dimensions Mapping}
\begin{tabular}{ll}
\toprule
\textbf{Index Range} & \textbf{State and Action Dimensions} \\
\midrule
0:6     & left qpos \\
6       & left gripper \\
7:13    & right qpos \\
13      & right gripper \\
14:17   & left endpose xyz \\
17:23   & left end pose $R$[:3,:2] \\
23:26   & right endpose xyz \\
26:32   & right end pose $R$[:3,:2] \\
\bottomrule
\end{tabular}
\label{table:dim}
\end{table}

\textbf{Obtaining the point cloud.} Given RGB-D observations from $N_v = 4$ calibrated viewpoints, we first project each pixel into 3D space using camera intrinsics $\mathbf{K}_i$ and extrinsics $[\mathbf{R}_i | \mathbf{t}_i]$:
\[
\mathbf{p}_k = \mathbf{R}_i^{-1} \left( \mathbf{K}_i^{-1} \begin{bmatrix} u_k \\ v_k \\ 1 \end{bmatrix} d_k - \mathbf{t}_i \right)
\]
where $\mathbf{p}_k \in \mathbb{R}^3$ denotes the 3D position of the $k$-th point, $(u_k, v_k)$ its pixel coordinates, and $d_k$ the measured depth. The concatenated point cloud $\mathcal{P} = \{ \mathbf{p}_k \}_{k=1}^M$ contains $M$ points, each represented as $\mathbf{p}_k = (x_k, y_k, z_k, r_k, g_k, b_k) \in \mathbb{R}^6$.

\noindent\textbf{Local geometric feature augmentation of point cloud.} For each point $\mathbf{p}_k$, we compute differential geometric features using its $K$-nearest neighborhood $\mathcal{N}_k = \{ \mathbf{p}_j \mid j \in \text{KNN}(k) \}$:
\begin{enumerate}
    \item \textit{Surface Normal $\mathbf{n}_k$}: Estimated via PCA on the covariance matrix:
    \[
    \mathbf{C}_k = \frac{1}{|\mathcal{N}_k|} \sum_{\mathbf{p}_j \in \mathcal{N}_k} (\mathbf{p}_j - \boldsymbol{\mu}_k)(\mathbf{p}_j - \boldsymbol{\mu}_k)^\top
    \]
    where $\boldsymbol{\mu}_k = \frac{1}{|\mathcal{N}_k|} \sum_{\mathbf{p}_j \in \mathcal{N}_k} \mathbf{p}_j$. The normal $\mathbf{n}_k$ corresponds to the eigenvector of $\mathbf{C}_k$ with the smallest eigenvalue.
    
    \item \textit{Curvature Features $\mathbf{c}_k$}: Computed by analyzing normal variations:
    \[
    \mathbf{V}_k = \left[ \mathbf{n}_k \times \mathbf{n}_j \mid \forall \mathbf{p}_j \in \mathcal{N}_k \right] \in \mathbb{R}^{3 \times K}
    \]
    The covariance $\boldsymbol{\Sigma}_k = \mathbf{V}_k\mathbf{V}_k^\top$ is decomposed to obtain eigenvalues $\lambda_1 \geq \lambda_2 \geq \lambda_3$. The curvature features are:
    \[
    \mathbf{c}_k = \left( \log(\lambda_1 + \epsilon),  \log(\lambda_2 + \epsilon) \right)
    \]
    where $\epsilon = 10^{-5}$ prevents numerical instability. 
\end{enumerate}

Finally, the augmented point representation is:
\[
\tilde{\mathbf{p}}_k = \left( x_k, y_k, z_k, r_k, g_k, b_k, \mathbf{n}_k, \hat{\mathbf{c}}_k \right) \in \mathbb{R}^{11}
\]

\noindent\textbf{Down sampling.} To handle computational constraints, we apply Farthest Point Sampling (FPS) to select $N = 4096$ points.
\[
\mathcal{S} = \text{FPS}(\mathcal{P}, N) = \underset{\mathcal{S} \subset \mathcal{P}}{\text{argmax}} \min_{\mathbf{p}_i, \mathbf{p}_j \in \mathcal{S}} \|\mathbf{p}_i - \mathbf{p}_j\|_2
\]
with $\mathcal{S}$ forming the final input to the network.

\begin{figure}
    \centering
    \hspace*{-0.04\linewidth} 
    \includegraphics[width=1.07\linewidth]{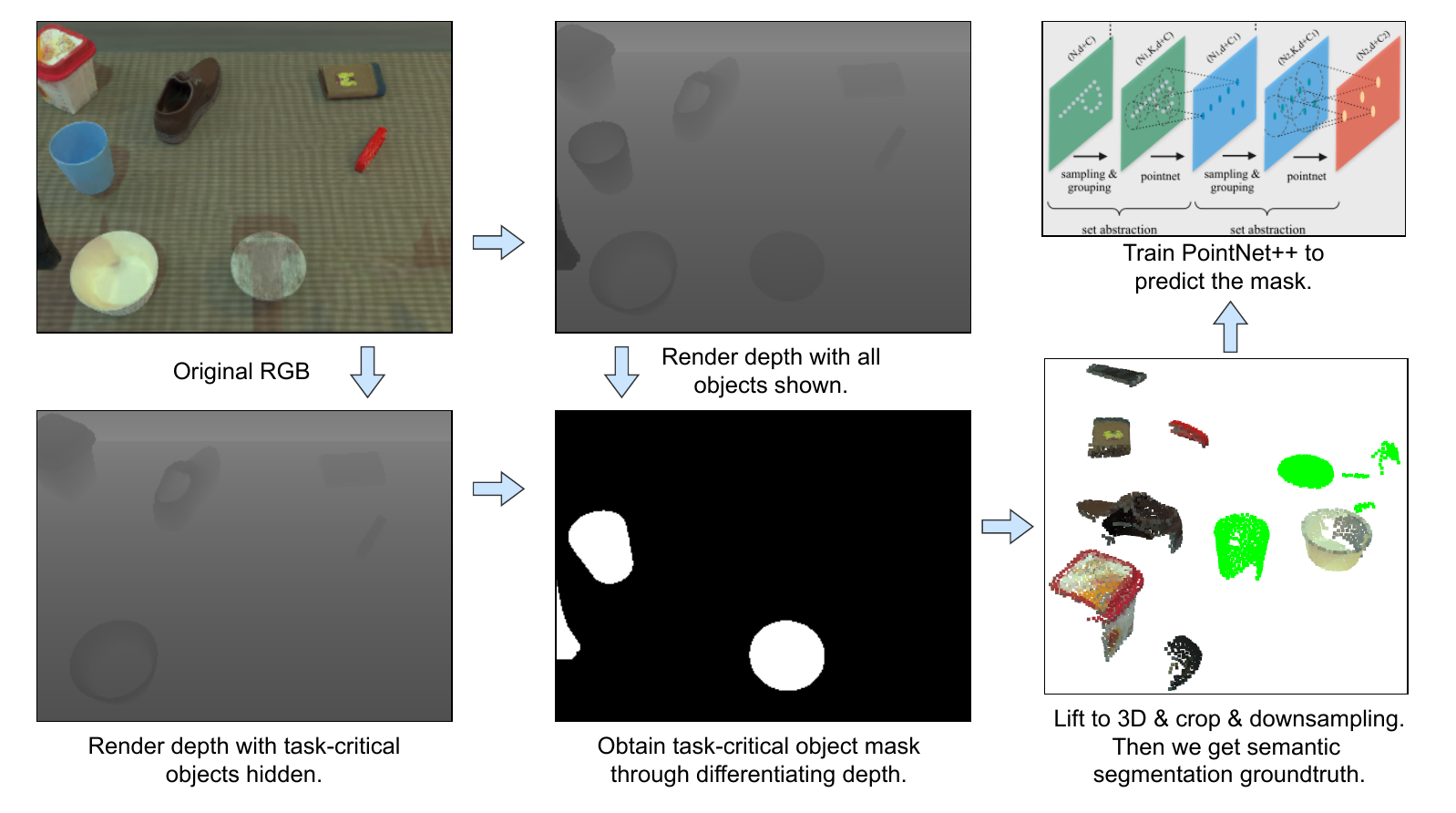}
    \caption{Simulator-Supervised Semantic Segmentation.}
    \label{fig:segment}
\end{figure}

\textbf{Simulator-supervised semantic segmentation.} To enhance the policy's focus on task-relevant objects in cluttered environments, we develop a simulator-supervised segmentation approach that automatically labels points belonging to critical objects, as shown in Figure~\ref{fig:segment}. The segmentation labels are generated directly within the simulation environment using a differential rendering technique:
\begin{enumerate}
    \item \textbf{Task-specific object list.} Before simulation starts, we define a list $\mathcal{O} = \{o_1, o_2, \dots, o_K\}$ of task-critical objects (e.g., blocks, phone, shoes) including the robot body.
    
    \item \textbf{Full scene rendering.} Render the complete scene to obtain RGB-D images from all $N_v$ viewpoints.
    
    \item \textbf{Occluded scene rendering.} For each viewpoint, re-render the scene while setting visibility=False for all objects in $\mathcal{O}$. This produces an \textit{occluded depth map} where critical objects are absent.
    
    \item \textbf{Mask generation.} Compute the depth difference:
    \[
    \Delta D(u,v) = D_{\text{occluded}}(u,v) - D_{\text{full}}(u,v)
    \]
    Pixels with $\Delta D(u,v) < \delta=10^{-5}$ (where $\delta$ is a small tolerance threshold) indicate locations where critical objects were present. This forms a 2D segmentation mask.
    
    \item \textbf{3D projection.} Project the 2D mask back to 3D space, assigning each 3D point a binary label $l_k \in \{0,1\}$ (1 for critical objects).
\end{enumerate}

This automated process generates precise point-level annotations without manual labeling, leveraging the simulator's complete scene knowledge. Next, we train a lightweight PointNet++ model to predict the critical object labels with the 11-dimensional downsampled point cloud $\mathcal{S}$ as inputs and per-point binary prediction $\hat{l}_k \in [0,1]$ as outputs.

\textbf{Language instruction following.} To address the competition requirement of executing diverse tasks based on natural language instructions, we developed a hierarchical conditioning system. This system comprises two core components: 1) a language instruction classifier that maps text to logits, and 2) a set of task-specific feature extractors that process observation data according to the identified task type. We use 8 distinct manipulation tasks based on object and action types. While the competition originally specified 6 task categories, we further divided the \texttt{place\_object\_scale} category into three separate tasks (\texttt{place\_mouse}, \texttt{place\_stapler}, \texttt{place\_bell}) due to significant shape variations between these objects. To train the classifier, we generated a dataset of 10,000 diverse language instructions through systematic paraphrasing and combinatorial augmentation of task descriptions. A pre-trained BERT-base model was then fine-tuned on this dataset.

\textbf{Task-conditioned observation encoders.} The policy employs eight parallel point cloud encoders $\{E_1, E_2, \dots, E_8\}$. Every encoder follows an identical lightweight architecture consisting of:
\begin{itemize} \small
    \item Per-point MLP (128-dimensional hidden layer)
    \item Max-pooling aggregation across points
    \item Final linear projection to 192-dimensional task embedding
\end{itemize}
During execution, the BERT classifier's output logits $\mathbf{l} \in \mathbb{R}^8$ determine the active encoder through an $\arg\max$ selection:
\[
k = \underset{i}{\arg\max} \, l_i
\]
Only the selected encoder $E_k$ processes the current point cloud observation $\mathcal{P}$, producing the task-conditioned feature vector $\mathbf{f}_k = E_k(\mathcal{P})$.

\textbf{Multi-task training protocol.} All encoders and the action expert are trained end-to-end using mixed-task batches containing trajectories from all 8 categories. For each trajectory, only the encoder corresponding to its task receives gradient updates. The shared diffusion action expert $A$ processes the feature vector $\mathbf{f}_k$ from the encoder and updates its parameters. Task-specific encoder weights remain isolated with no parameter sharing. This approach enables efficient knowledge transfer to the shared action model while preventing negative interference between distinct manipulation strategies.

\textbf{Architectural simplicity justification.} Consistent with findings in 3D Diffusion Policy (DP3) literature, we observed that complex point cloud networks (e.g., PointNet++, PointNeXt) significantly hindered training stability. The minimalist MLP architecture provided the best convergence behavior compared to hierarchical point processors. This design choice prioritizes robust gradient flow and computational efficiency-critical factors given the 24GB GPU memory constraint.

\subsection{Affordance-Anchored Keypose Diffusion}

Conventional diffusion policies trained on dense action sequences (typically sampled at 20-25Hz) exhibit inherent limitations in robotic manipulation. We observed that most trajectory actions constitute \textit{low-entropy motions} – predominantly inertial movements where future states can be extrapolated from current kinematic conditions. Training on the dense action datasets induces two critical issues:

\begin{itemize}
    \item \textbf{Erroneous causality learning.} Policies learn spurious correlations (e.g., "I move forward because I was moving forward in the past observation.") rather than goal-oriented affordance grounding ("I move forward because of the target object.")
    \item \textbf{Decision dilution.} Computational resources are wasted on trivial predictions while critical transitions remain under-represented.
\end{itemize}

\textbf{Action space reformulation.} Existing solutions like action chunking partially mitigate these issues by extending prediction horizons but fail to address the fundamental sparsity of meaningful decisions. Mirroring human motor control – where conscious planning occurs only at kinematic inflection points, such as approach initiation and contact establishment, while transit phases remain subconscious – we reformulate action prediction exclusively around geometrically significant keyposes.

\textbf{Affordance-Anchored Keyposes.} We define keyposes as affordance-grounded kinematic states marking transitions between motion primitives. For common grasping and placing manipulation skills, the keyposes can be marked as:
\begin{itemize}
\small
    \item {Pre-Grasp (PG):} End-effector positioned at affordance-aligned approach vector.
    \item {Target Grasp Open (TGO):} Gripper centered on grasp affordance with maximum aperture.
    \item {Target Grasp Closed (TGC):} Affordance-optimized grasp closure pose.
    \item {Pre-Place (PP):} Object transported to placement approach vector.
    \item {Target Place Closed (TPC):} Object positioned at target affordance.
    \item {Target Place Open (TPO):} Gripper retraction after placement.
    \item {Home (HM):} Neutral configuration.
\end{itemize}

Crucially, these poses are geometrically anchored to object affordances through linear transformations, e.g., PG = affordance centroid + pre-grasp offset. Therefore, the keyposes are also denoted as \textit{\textbf{action anchors}}. A grasp-place trajectory can be represented as PG-TGO-TGC-PG-PP-TPC-TPO-PP-HM in chronological order. For other robotic manipulation skills, they can also be divided according to similar principles. The reformation of action space creates a compressed action space where the policy only needs to learn 10-30 sparse action anchors per task rather than thousands of dense actions. The neural networks do not explicitly learn the names and tags of each action anchor, but implicitly learn to generate a list of multiple future action anchors through end-to-end diffusion training.

\textbf{Anchor tagging.} Leveraging RoboTwin's modular expert policy structure, the atomic operation boundaries, e.g., \texttt{MovetoPose}, \texttt{SetGripper} explicitly define keypose transitions. The terminal action of each atomic module is tagged as a keypose. For non-symbolic human demonstrations, the key poses can be detected via kinematic discontinuity because slowing down and re-accelerating when reaching a pose that requires a change in motion state is in line with human operating habits.

\textbf{Diffusion-based action expert.} We use the Conditional U-Net 1D network as the action expert \cite{diffusionpolicy,diffusionpolicy3d}. One action expert network is adopted for all the tasks. The Feature-wise Linear Modulation (FiLM) conditioning method \cite{filmcondition} is adopted to integrate the features extracted from observations. The action expert uses only one frame of current observation as input, with no historical information. Our diffusion U-Net predicts action anchor sequences $[a_{A_{k+1}}, ..., a_{A_{k+H}}]$ where each action anchor $a_{A_{k}} \in \mathbb{R}^{32}$ represents joint angles and end-effector pose. During training, all observation states are supervised to predict only the next $H$ keyposes, where $H$ is the prediction horizon. We use $H=8$ in AnchorDP3. However, only the first predicted action anchor is executed during inference.

\textbf{Full-state supervised training.} We use a simple and unified L2 loss for all the 8$\times$32 output elements,  where 32 is the augmented action dimension and 8 is the prediction horizon.  Although the only useful output is located at the top left corner 1$\times$14 because only the first action anchor is executed without referring to end poses, the full state supervision for all the elements is beneficial for converging.

\subsection{Large-scale dataset construction based on action anchors}

\begin{figure}
    \centering
    \hspace*{-0.04\linewidth} 
    \includegraphics[width=1.07\linewidth]{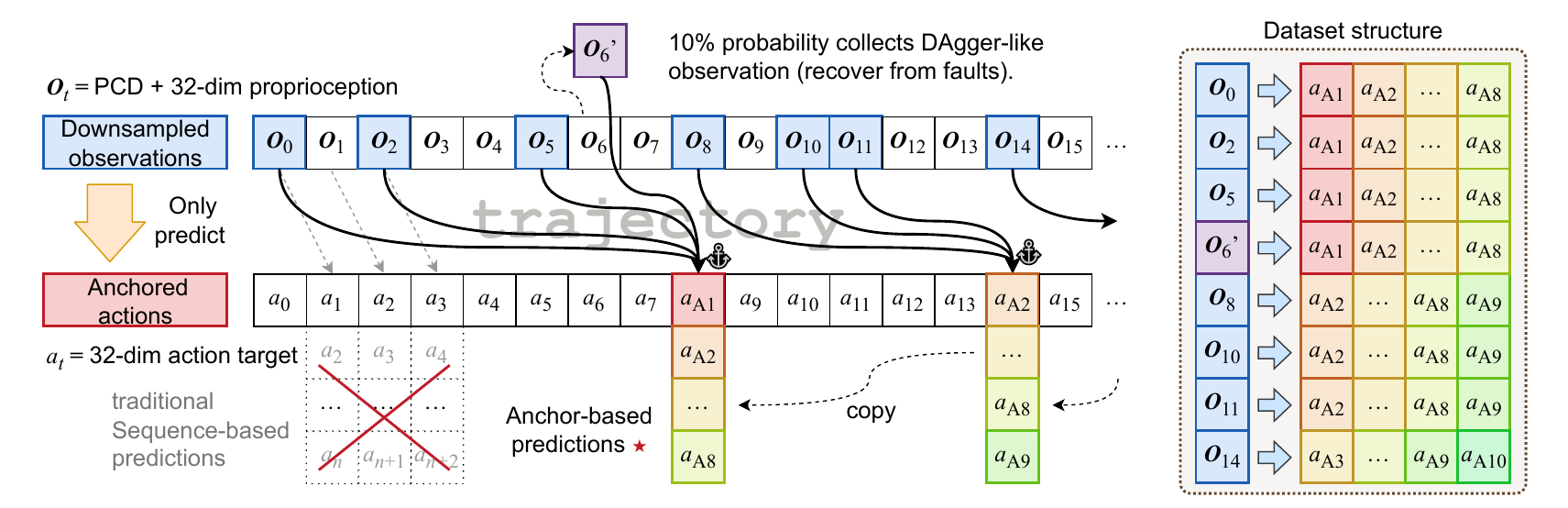}
    \caption{RoboTwin dataset construction based on action anchors.}
    \label{fig:dataset}
\end{figure}

Conventional dataset construction for robotic policies captures dense state-action pairs at full control frequency (typically 20-25Hz). We observed this approach incurs significant redundancy, exhibiting kinematic monotonicity where actions are predictable through simple inertial extrapolation. Beyond computational waste, this redundancy: 1) Dilutes the representation of critical kinematic inflection points; 2) Limits environmental diversity due to rendering/storage bottlenecks; 3) Encourages policies to learn motion inertia rather than affordance-driven decisions. To overcome these limitations, we implement a resource-optimized sparse trajectory collection protocol, as shown in Figure~\ref{fig:dataset}. While the total data volume remains similar, our proposed method can significantly reduce data frames per trajectory, resulting in a larger number of individual trajectories.

\textbf{Two-phase data collection.} Our data collection strategy for each trajectory is divided into two distinct phases: a rollout phase and a render phase. The rollout phase is executed solely on the CPU, running the expert policy without rendering. This phase serves a dual purpose: it rapidly filters out configurations that are impossible to complete the task, such as objects not being on the table, and, for successfully completed tasks, records the specific time steps where actions are designated as action anchors. Then, the render phase constructs the sparse trajectory dataset using the proposed strategy.

\textbf{Sparse trajectory dataset construction.} Within a single trajectory, $\mathbf{O}_t$ represents the observation, including point cloud data and 32-dimensional proprioception, while $a_t$ denotes the control target under the current observation, also 32-dimensional. Instead of the traditional sequence-based method that saves all $a_t$, our method only saves the action anchors $a_{Ak}$. Each frame of observation is directly pointed to a list of future action anchors $[a_{A(k+1)},...,a_{A(k+n)}]$. Under this strategy, it can be observed that rendering every observation frame between two key action time points is inefficient because their labels are identical. Therefore, we employ a random sampling strategy, rendering only two randomly chosen frames between the two consecutive key action time points. By reducing the identical-labeled observations, the size of a trajectory is significantly reduced. The number of trajectories with diverse environments can be \textbf{increased by 14.5 times} within the same computational budget and dataset size. The increased exposure to different environmental settings contributes to the improvement of the model’s accuracy.

\textbf{Data aggregation (DAgger).} During the render phase, there is a 10\% probability of collecting observations from outside of the expert demonstration trajectories. We manually insert code snippets into the expert demonstrations that command the system to go to a wrong pose and obtain a frame of observation. These DAgger observations typically deviate from the optimal trajectory or are about to engage with incorrect affordances. Therefore, the policy is trained to predict actions back to correct action anchors from the erroneous observations. By including examples of recovery from suboptimal states, the robustness of the resulting policy is improved.

\section{Conclusion}
This work addresses the critical challenges of dual-arm robotic manipulation in unstructured environments through AnchorDP3, a novel diffusion policy framework that achieved state-of-the-art performance in the RoboTwin Challenge. Our solution directly tackles three fundamental limitations identified during the competition. First, \textbf{perceptual ambiguity in cluttered scenes} is resolved through our \textit{simulator-supervised semantic segmentation} module, which leverages rendered ground truth to inject affordance awareness into point cloud representations. Second, \textbf{multi-task interference} is mitigated via \textit{task-conditioned feature encoders}, enabling specialized feature extraction while maintaining a unified action expert. Third, \textbf{inefficient long-horizon action modeling} is overcome by our \textit{affordance-anchored keypose diffusion} approach, which predicts geometrically meaningful sparse keyposes supervised with full kinematic state. These innovations collectively enabled AnchorDP3 to achieve a 98.7\% average success rate under extreme scene randomization - the highest result in the RoboTwin Challenge. By seamlessly integrating with procedurally generated simulation workflows, AnchorDP3 provides a compelling pathway toward tackling complex manipulation tasks. Future work will explore sim-to-real transfer and extension to dynamic environments.

\clearpage
{
\small
\bibliographystyle{unsrt}
\bibliography{ref}
}

\end{document}